\documentclass[lettersize,journal]{IEEEtran}

\usepackage{amsmath,amsfonts}
\usepackage{algorithmic}
\usepackage{algorithm}
\usepackage{array}
\usepackage[caption=false,font=normalsize,labelfont=sf,textfont=sf]{subfig}
\usepackage{textcomp}
\usepackage{stfloats}
\usepackage{url}
\usepackage{verbatim}
\usepackage{graphicx}
\usepackage{soul}
\usepackage{xcolor}
\usepackage{amsmath,amsfonts} 

\DeclareMathOperator*{\argmaxA}{arg\,max}
\usepackage{subfig}
\usepackage{threeparttable}
\usepackage{graphicx}         
\usepackage{multirow}
\usepackage{array}    
\usepackage[implicit=fals,hidelinks]{hyperref}
\usepackage[backend=bibtex,style=ieee,natbib=true,sorting=none,url=false,doi=false,isbn=false]{biblatex} 
\AtBeginBibliography{\small}
\newcommand*{\rom}[1]{\expandafter\@slowromancap\romannumeral #1@}

\begin{document}

\title{Open-Set Object Recognition Using Mechanical Properties During Interaction}

\author{Pakorn Uttayopas, Xiaoxiao Cheng and Etienne Burdet


\thanks{Department of Bioengineering, Imperial College of Science, Technology and Medicine, London, UK. \{pu18, xcheng4, e.burdet\}@imperial.ac.uk.}

\thanks{\textit{Corresponding authors: Pakorn Uttayopas, Xiaoxiao Cheng.}}
}

\markboth{Journal of \LaTeX\ Class Files,~Vol.~14, No.~8, August~2021}%
{Shell \MakeLowercase{\textit{et al.}}: A Sample Article Using IEEEtran.cls for IEEE Journals}

\IEEEoverridecommandlockouts
\IEEEpubid{\begin{minipage}[t]{\textwidth}\ \\[10pt]
        \centering\normalsize{0000--0000/00\$00.00~\copyright~2021 IEEE}
\end{minipage}}

\maketitle

\begin{abstract}
Current robotic haptic object recognition...
\end{abstract}

\begin{IEEEkeywords}
open-set setting, haptic exploration, supervised learning for classification, unsupervised learning for clustering.
\end{IEEEkeywords}

\section{Introduction}
Robots become more common to operate in household environments surrounded by various kinds of objects. When vision is unavailable, tactile perception helps robots interact with such an environment more effectively \cite{Luo2017Mechatronics}. However, most of the existing work was done in a close-set condition in which testing objects always belong to one of the objects in the training set \cite{Dallaire2014, Bhattacharjee2018, Taunyazov2019, Cao2020}. This is different from the open-set condition where testing objects can be beyond the knowledge of the robots. This may cause robots to fail in recognising objects and, consequently, in object manipulation.

Humans, on the other hand, can easily handle novel objects in open-set conditions by learning and gathering new information from the exploration process and updating their knowledge. For robots to handle such a situation, a concept of open-set recognition (OSR) has been proposed as the first step for robots to interact with novel objects \cite{Abhijit2015}. Unlike traditional close-set recognition, OSR enables robots to classify the samples of known objects and also detect samples of novel objects which are collected and investigated to gain further information.

Previous works show that OSR has been successfully applied in various fields such as object detection \cite{Zheng2022, Wu2022, Ma2022}, face recognition \cite{Vareto2017, Gunther2017, Salomon2020}, and antonymous driving \cite{Bogdoll2022}. However, these systems are not able to incrementally learn new information from novel objects and label them. This could be addressed by involving humans to assign specific labels to novel objects or by implementing semi-supervised learning in the recognition system for incremental automatic labeling \cite{cao22}. The simplest method might be to apply unsupervised learning methods after finishing the novelty detection \cite{Joseph2021}.

While OSR has been widely applied in the field of vision \cite{Boult2019}, it is rarely seen in tactile-based applications. Liu et al. \cite{Liu2023} introduced, for the first time, the robotic material recognition system for open-set conditions. The framework deployed a Generative Adversarial Network (GAN) to generate fake novel tactile images as an example used in prototype-based learning \cite{Yang2018} to prepare for novelty detection. However, the system does not specifically label the detected novel objects and requires a lot of tactile images to learn. Besides, the values of tactile information can be affected by interactions, causing confusion among materials.

Unlike tactile information, mechanical properties such as coefficient of restitution, viscoelasticity, and friction coefficient provide a unique representation of objects \cite{Su2015, Ren2017, Stronge2018} and are well distinguishable among objects \cite{ Bednarek2021, Yane2022, Uttayopas2023}. Inspired by these properties, we proposed a simple open-set tactile recognition framework that uses objects' mechanical properties to perform classification for known objects and clustering for novel objects. The classification can be done by a supervisor learning method while a distance-based unsupervised learning method is used for clustering.

Since the mechanical properties of novel objects cannot be observed during training, we exploit knowledge of known objects to estimate the new cluster's centre and size via a regression model, as this approach was used to generate samples in zero-shot learning methods \cite{Verma2017}. This is intended to enable the framework to cluster novel objects online, without a complete set of samples. Furthermore, it also aims to make the new cluster as close as the ground truth cluster, unlike general clustering algorithms where the initial cluster's centre is randomly selected.


In this paper, we extend the framework in \cite{Uttayopas2023} that estimates mechanical protest during interaction to be able to perform open-set recognition and incrementally label novel objects. The framework is validated using the estimated mechanical properties collected in \cite{Uttayopas2023}. Our results are compared with alternative methods used to deal with open-set conditions. Lastly, the role of the framework's hyperparameter on novel object clustering results is investigated.

\begin{figure}[tb]
\centering
\qquad \qquad
\includegraphics[width=1\columnwidth]{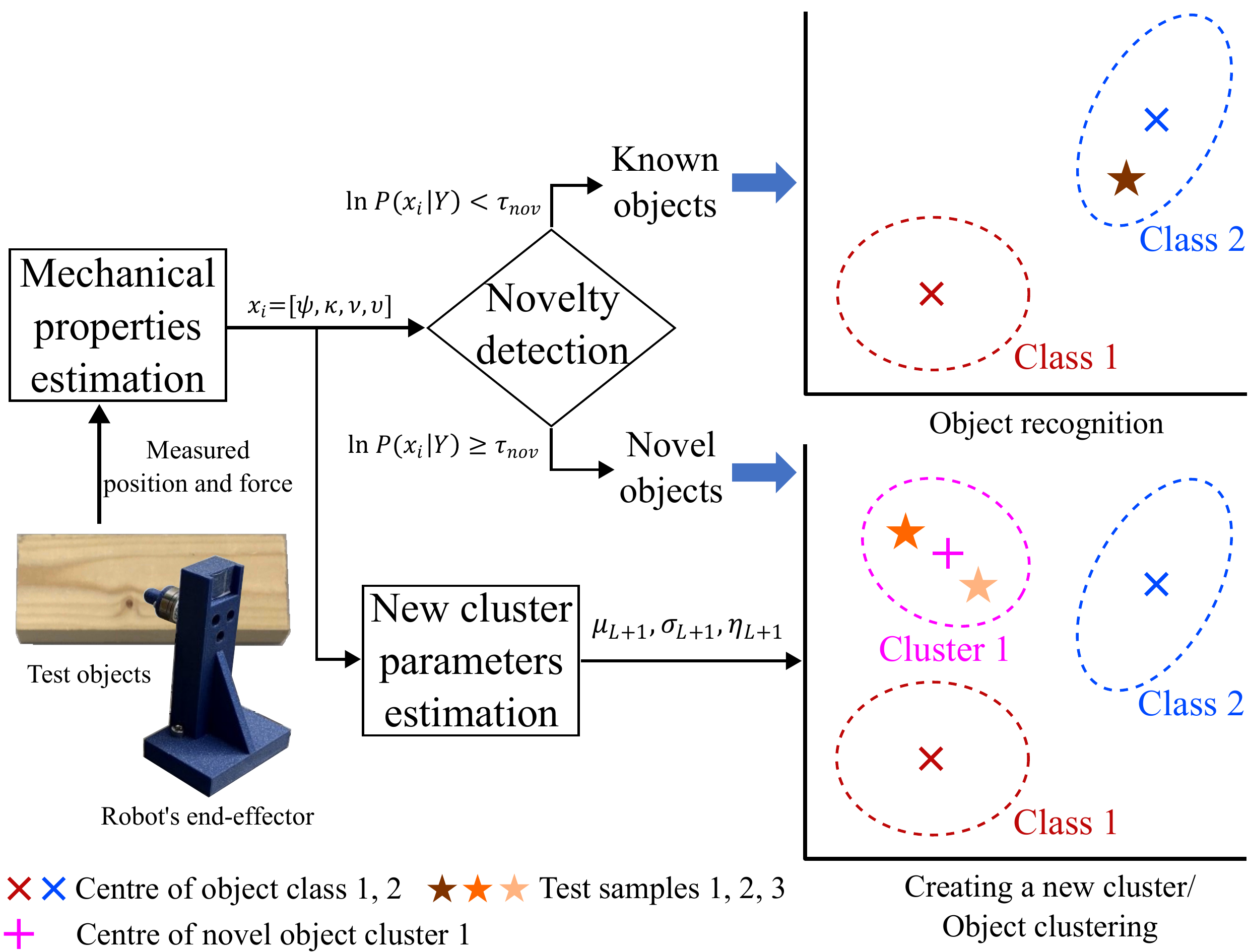}
\caption{Object recognition and clustering process. The end-effector force and position measured during the interaction are used to obtain mechanical properties with an estimator. The estimated mechanical features are then used as a test sample $x_i$ to identify as either known or novel objects. Test sample 1 is one of the known objects, thus the classification is performed. Test sample 2 is a novel object, thus cluster 1 is generated by estimating its parameters. However, test sample 3 is also a novel object, but it is labeled as cluster 1 because it lies in the cluster boundary.}
\label{fig:diagram}
\end{figure}

\section{Object Recognition Framework}
\subsection{Problem Statement}
\label{sec:problem}
In the problem that we investigate, a robot is trained to recognise the set of objects class $Y=\{y_1,y_2,..., N\}$ from mechanical properties and test it on the set $O=Y\cup Z$ containing the known classes $Y$ and novel classes $Z$ (unlabelled), s.t. $Y\cap Z= \emptyset$ and $Y\neq Z$. T. During the training phase, a robot learns parameters from a training dataset $D_{train}=\{(y_i,(x_1,...x_{TN})\}_{i=1}^N$, where $TN$ is the number of the training sample and $x$ represents the mechanical properties collected from a certain space $X$. During the test phase, a test sample $x\in X$ is either classified as one of the known classes or discovered as a novel class and clustered accordingly. This allows the number of clusters of novel objects to grow as a new class is discovered $U=\{u_1,u_2,..., L\}$. We assume that the robot does not have any prior knowledge of whether the test sample belongs to known classes or unlabelled classes and only knows the number of known classes. 


%

\subsection{Framework Overview}
Fig.\,\ref{fig:diagram} shows the proposed framework with its two components; mechanical properties estimation, and recognition$\slash$\, clustering. A robot interacts haptically with a test object to retrieve interaction force at different positions to estimate an object's mechanical properties such as coefficient of restitution $\psi$, stiffness $\kappa$, viscosity $\nu$, and friction coefficient $\upsilon$ via the dual Kalman filter technique \cite{Uttayopas2023}. These estimated mechanical properties are used as a test sample $x_i$ to either recognise the known objects or label them as novel objects depending on the novelty detection. 

The novelty detection is based on the probability density function of $x_i$ given the training mechanical properties. This reflects how likely $x_i$ belongs to one of the known object classes. Specifically, we calculate a log-likelihood of $x_i$ across all the known classes $\ln p(x_i|Y)$ based on classes' Gaussian parameters. We determine whether the test sample belongs to one of the known classes or the novel class as follows
\begin{equation}
c = 
\begin{cases} \argmaxA_{y\in Y}p(y|x_i)\,, \quad  \ln p(x_i|Y) < \tau_{nov} & \\
novel\,, \qquad  \qquad \qquad  \,\,\,\,\,  \text{otherwise} &
\end{cases}
\end{equation}
where $\tau_{nov}$ is a novelty threshold, and $p(y|x_i)$ is the posterior of a known object’s class $y$ given $x_i$ calculated from the Naive Bayes classifier.

For novel classes, a distance-based clustering method is used to decide which cluster of $x_i$ belongs. In brief, $x_i$ is labeled as the same class as the closest existing cluster that $x_i$ lies within its boundary $\eta_c$. If $x$ does not lie in any of the existing cluster boundaries, the new cluster is created. This is done by estimating parameters describing the new cluster such as the centre point and its boundary via a regression model from the training dataset (details in section \ref{sec:cluster algorithm}).

Since novel clusters are synthetically created, their parameters need to be updated with the real sample. This is done once the number of datapoints within clusters reaches a particular number $\tau_{update}$. After all datapoints are fed, sample points that lie within the cluster having the number of datapoints lower than $\tau_{out}$ are treated as outliners and removed from the framework. 
\section{Clustering Algorithm}
\label{sec:cluster algorithm}
\subsection{Labelling the test sample}
As $x_i$ is identified as a novel object, it could belong to the existing cluster of novel objects. The probability of $x_i$ belonging to an existing cluster is calculated based on a distance
\begin{equation}
p(u|x_i)=\frac{\exp(-\frac{1}{2}\,d_M(x_i,\mu_u))}{\sum_{u'=1}^{L}\exp(-\frac{1}{2}\,d_M(x_i,\mu_{u'}))}
\label{eq:cluster_results}
\end{equation}
where $\mu_u$ is mean values of each mechanical properties in cluster $u$, and $d_M(x_i,\mu_u)$ is the Mahalanobis distance between $x_i$ and a cluster centre $\mu_u$ considering the cluster covariance $ \Sigma_u$
\begin{equation}
d_M(x_i,\mu_u) =(x_i-\mu_u)^\top \Sigma_u^{-1}(x_i-\mu_u).
\label{eq:distance}
\end{equation}
The label could be assigned as 
\begin{equation}
c =\argmaxA_{u\in U} p(u|x_i).
\end{equation}

However, if $x_i$ lies beyond the boundary of the most probable cluster, it is discovered as a sample in a new cluster. The new cluster is created and the number of clusters is increased by 1. The final assigned label is given by 
\begin{equation}
c = 
\begin{cases} c\,, \qquad \quad \quad \quad d_M(x_i,\mu_{c})\leq \eta_c & \\
N+L+1\,, \quad  \text{otherwise} &
\end{cases}
\end{equation}
where $\eta_c$ is the boundary of the most probable cluster $c$, the longest Mahalanobis distance between its points and center.

\subsection{Learning the regression model}
As a new cluster is discovered, a regression model is exploited for its ability to predict the cluster parameters without any knowledge of the novel object. The relationship between the feature vector $a_y$ and its cluster parameters $\theta_y$ for each class $y$ can be defined as $\theta_y=\,f_{\theta}(a_y)$ where $f_{\theta}$ is a mapping function and could consist of multiple functions depending on the number of parameters in $\theta_y$. 

We define $\theta_y$ as the mean vector $\mu_y\in\mathbb{R}^D$ and covariance matrix $\Sigma_y\in\mathbb{R}^{D\times D}$ in $D$ dimensional Gaussian distribution and assume that $\Sigma_y=diag(\sigma_y^2)$ where $\sigma_y^2=[\sigma_{y,1}^2,...,\sigma_{y,D}^2]$ for simplification. The parameters $\mu_y$ and $\sigma_y^2$ can be modelled as linear functions of mechanical properties vector $a_y$:

\begin{equation}
\mu_y \,= \mathbf{W}_{\mu}\,a_y, \qquad y=1,...,N
\end{equation}
\begin{equation}
\sigma_y^2 \,= \mathbf{W}_{\sigma^2}\,a_y, \qquad y=1,...,N
\end{equation}
where $\mathbf{W}\in\mathbb{R}^{D\times K}$ is the output weight matrix and its subscript indicates which parameters it belongs to. 

The least-squares technique with Tikhonov regularisation is applied to match an estimation of $\mu_y$ and $\sigma_y^2$ with the training data. We denote $\mathbf{M}=[\mu_1,...,\mu_M]\in\mathbb{R}^{D\times M}$, $\mathbf{R}=[\sigma^2_1,...,\sigma^2_M]\in\mathbb{R}^{D\times M}$, $\mathbf{A}=[a_1,...,a_M]\in\mathbb{R}^{K\times M}$, thus $\mathbf{W}$ is estimated as follows 

\begin{equation}
\hat{\mathbf{W}}_{\mu} = \mathbf{M}\mathbf{A}^\top(\mathbf{A}\mathbf{A}^\top+\lambda_{\mu}\mathbf{I})^{-1}
\end{equation}
\begin{equation}
\hat{\mathbf{W}}_{\sigma^2} = \mathbf{R}\mathbf{A}^\top(\mathbf{A}\mathbf{A}^\top+\lambda_{\sigma^2}\mathbf{I})^{-1}
\end{equation}
where $\lambda$ is the regularisation pa rameter which is used to prevent overfitting to the training data, and $\mathbf{I}$ is the identity matrix.

Inspiring by the next-generation reservoir computing \cite{Gauthier2021}, the feature vector is composed of a linear and nonlinear part of the feature $a_y=a_{lin,y}\oplus a_{nonlin,y}$ where $\oplus$ is the vector concentration. The mechanical properties are the main attributes used in the framework, the linear feature is $a_{lin,y}=[\kappa, \nu, \psi, \upsilon]^\top$ and the quadratic polynomial function was applied on $a_{lin,y}$ to obtain nonlinear part of feature vector as $a_{nonlin,y}=[\kappa^2,\kappa\cdot\nu,..., \nu^2,\nu\cdot\psi,..., \psi\cdot\upsilon, \upsilon^2]^\top$. Totalling, $a_{y}$ contains 14 features and each known object class has $TN$ datapoints, the feature vector becomes $a_y=[a_{y,1},..., a_{y,TN}]$, thus $D=4$, $K=14$ and $M=N\cdot TN$.

\subsection{Creating a new cluster}

Given the estimated weight $\hat{\mathbf{W}}_{\mu}$ and $\hat{\mathbf{W}}_{\sigma^2}$, parameters describing the new cluster can be calculated as follows
\begin{equation}
\mu_{L+1} \,= (1-\alpha)\,\hat{\mathbf{W}}_{\mu}a_{x_i}+\alpha\,x_i
\end{equation}
\begin{equation}
\sigma_{L+1}^2 \,= \beta\,\hat{\mathbf{W}}_{\sigma^2}\,a_{x_i}
\end{equation}
where $\alpha\in [0,1]$ is a hyperparameter indicating centre position ($\alpha=0$: the predicted mean as centre and $\alpha=1$: the test sample $x_i$ as centre) and $\beta\in [0,\infty)$ is a hyperparamter indicating boundary of cluster (higher $\beta$ results in larger boundary). 

To define cluster boundary, eq.\, (\ref{eq:distance}) is calculated based on $N_{gen}$ points of generated datapoints $X_{gen} \sim \mathcal{N}(\mu,\,\sigma^{2})$ to search for the longest distance from $X_{gen}$ to the centre. All parameters of the cluster are updated once the number of datapoints reaches the threshold $\tau_{update}$. Note that $X_{gen}$ is included to update the parameters with the test datapoints.

\section{Experimental Setup}
\subsection{Data collecting}
\begin{figure}[!bt]
\centering
\qquad \qquad
\includegraphics[width=1\columnwidth]{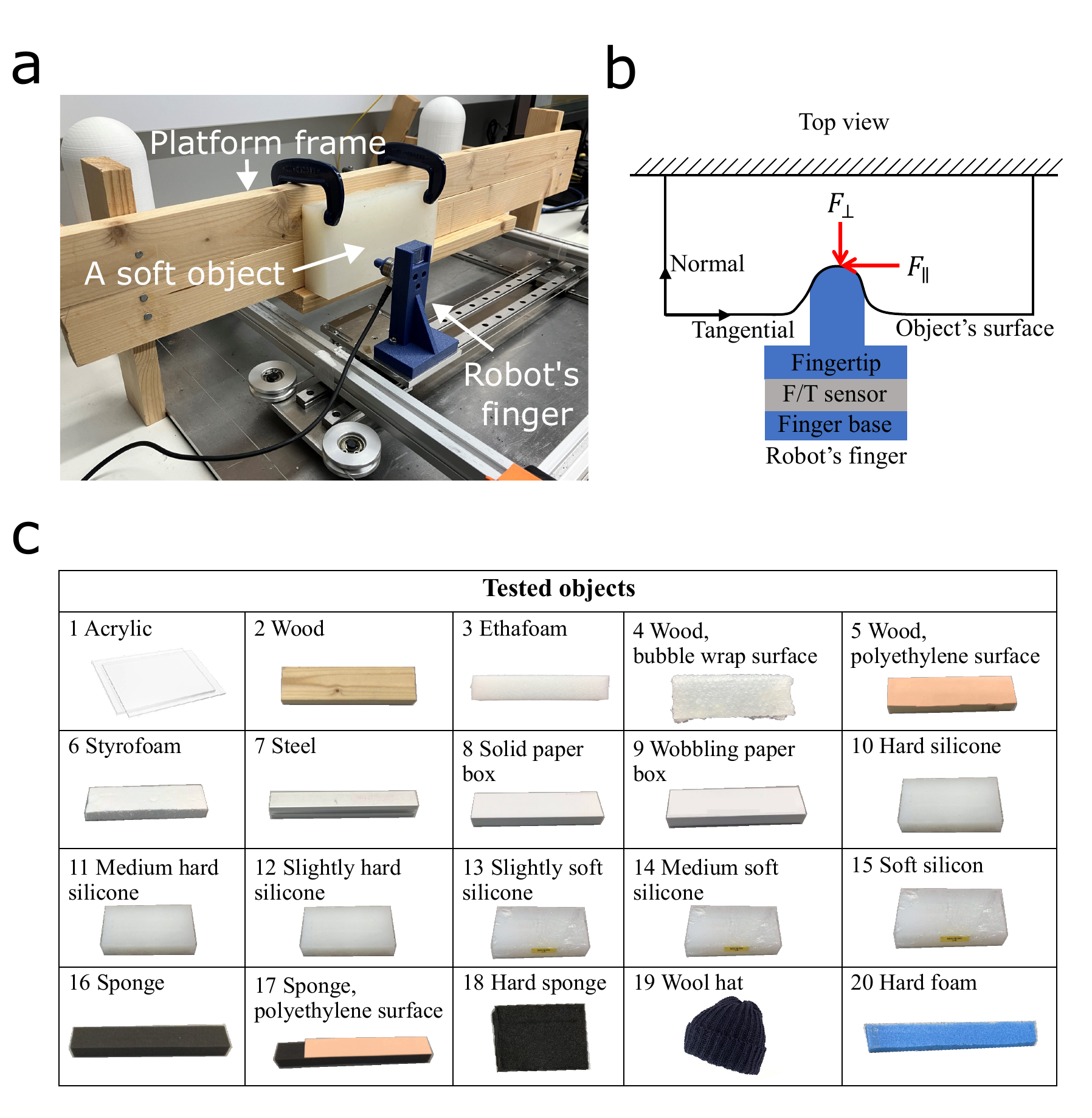}

\caption{Experimental setup. (a) HMan robot with a sensorized finger and an object to examine. A wooden platform frame is used to attach various objects for the robot to explore. The finger is driven by two motors in the normal and tangential directions to the object surface. (b) Diagram of robot finger interacting with an object's surface. (c) 20 objects used in the experiment.}
\label{fig:H-man}
\end{figure}

To test the developed object recognition and clustering framework, it was applied to mechanical properties data collected in \cite{Uttayopas2023}. The HMan robot \cite{Campolo2014} (shown in Fig.\,\ref{fig:H-man}a) was used to estimate object mechanical properties through robot-object interactions. This robot is a 2-DOF cable-driven planar robot controlled by two actuators (model 352699; Maxon Motor). These actuators were operated by the NI real-time system with a 1 kHz sampling rate. A finger equipped with a six-axis force sensor (SI-25-0.25; ATI Industrial Automation) was installed on the base of the robot's end-effector to measure forces during robot-object interaction with a 1 kHz sampling rate. 

The mechanical properties estimator \cite{Uttayopas2023} was implemented on the robot to estimate the coefficient of restitution, stiffness, viscosity, and friction coefficient of 20 objects (see Fig.\,\ref{fig:H-man}c). The estimation was done with interactions as follows 
\begin{itemize}
\item \textit{Tapping}:  
To estimate the coefficient of restitution, a constant force was exerted on the robot's end-effector for 0.5 seconds followed by its free movement toward the object's surface to create an impulse.

\item \textit{Indentation}: To estimate the surface viscoelasticity, the robot pressed its finger against the object's surface in the normal direction with a desired trajectory $r_{\perp}=0.01\sin(8t)+0.01\,m$, $t \in (0,20]\,s$.

\item \textit{Sliding}: To estimate the coefficient of friction, the robot moved its finger in the tangential direction on the object's surface with a constant velocity of 0.04\,m/s while maintaining a constant normal force of 4\,N.
\end{itemize}

These interactions were performed separately which allowed the robot to interact with objects smoothly and to have sufficient contact with the objects' surfaces. The estimation was repeated 25 times for each pair of actions and object.

\subsection{Data processing}
A dataset was formed based on the estimated mechanical properties. The coefficient of restitution was directly used as one of the features. The mean values of stiffness, viscosity, and friction coefficient for the last two seconds of estimation were extracted as the steady-state values and used as additional features. 

We assumed that the robot does not have any knowledge of some object classes, thereby the dataset was split into two sets of classes; known classes and novel classes. We randomly selected 12 classes (60\% of the number of classes) to be known classes and the remaining 8 classes (40\% of the number of classes) to be novel classes. 3/4 datapoints in the known class were randomly selected and used to learn object recognition parameters and regression model weights. On the other hand, the remaining 1/4 datapoints were used as the testing set combined with the set of novel classes. To ensure the robustness of the framework, the evaluation was done with 25 repetitions with different choices of known classes and novel classes each. 

\subsection{Alternative algorithms}

Given the problem stated in section \ref{sec:problem}, there are rarely ready-to-use baselines, especially in the context of haptic, that we can compare with our framework. Thus, we extended the existing novelty detection, supervised learning, and unsupervised learning methods as alternative methods to compare with our results in three tasks: 1.) how well it can distinguish between known and novel objects (novelty detection), 2.) how well it can recognise the known object (known object classification), and 3.) how well it can label novel objects (novel object clustering). 

In the first task, we compare our novelty detection results with three alternative methods: \{Isolated forest, Gaussian Mixture Models, and a Nearest Non-Outlier algorithm (NNO) \cite{Abhijit2015}\}. An isolated forest algorithm is similar to a random forest algorithm, but it is used to detect anomalies by searching for boundaries between anomalies and normal datapoints. GMM was proposed as a novelty detector in \cite{Abderrahmane2020}. It detected anomalies by comparing the log-likelihood of the fitted GMM given the test sample to a threshold. If the value is lower than the threshold, the test sample is novel. Lastly, the NNO is a distance-based classifier with the ability to detect anomalies. A distance function of a test sample. If the distance function value is lower than 0, the test sample is novel. 

In the second task, the test datapoints that are identified as known objects by the isolated forest and GMM will be classified using the same Naive Bayes classifier (NB) as our framework. On the other hand, the identified test datapoints as known objects from the NNO are classified by the nearest class mean classifier (NCM) \cite{Mensink2013} as the NMC is already integrated with the NNO algorithm. Lastly, we replace our classifier in the framework with another two alternative classifiers \{a fully connected neural network (NN) and a polynomial-kernel support vector machine (SVM)\} to compare classification results from them.

In the third task, four alternative clustering algorithms \{K-means, Incremental online k-means (IOKM) \cite{ABERNATHY2022}, DBscan, and DenStream \cite{Cao2006}\} are used to label test datapoints identified as novel objects by our novelty detector. Both K-means and IOKM are centroid-based clustering methods, but the latter is an online version of K-means. On the other hand, DBscan is a density-based clustering method that clusters areas with high density separately from low-density areas. It does not require a pre-defined number of clusters, unlike the first two algorithms. DenStream is the extension of DBscan that could work online. All alternative algorithms do not have the estimators to predict cluster centres and boundaries. 

We also compare our new cluster parameters estimator with two alternative methods: \{random and variational autoencoder (VAE)\}. For random, the new cluster centre and boundary are randomly generated. On the other hand, VAE is used to generate cloud points of novel objects based on the test datapoints. These cloud points then are used to calculate the new cluster centre and boundary. 

\subsection{Evaluation metrics}

The accuracy of known/novel object distinction in the first task and the known object recognition rate in the second task can be reported as the proportion of the number of datapoints recognised or identified correctly and the total number of datapoints. 

To evaluate clustering results in the third task, we compared the results of assigned labels (clustering results) of the novel objects with the true labels using the adjusted rand index (ARI):


\begin{equation}
\label{eq:ARI}
\text{ARI} = \frac{\sum_{ij}\binom{n_{ij}}{2} - \left[\sum_i\binom{a_i}{2}\sum_j\binom{b_j}{2}\right]/\binom{n}{2}}{\frac{1}{2}\left[\sum_i\binom{a_i}{2} + \sum_j\binom{b_j}{2}\right] - \left[\sum_i\binom{a_i}{2}\sum_j\binom{b_j}{2}\right]/\binom{n}{2}}
\end{equation}
where $n_{ij}$ is the number of datapoints in a cluster of known label $A_i$ and a cluster of assigned label $B_j$, $a_i$ and $b_j$ is the number of datapoints in cluster $A_i$ and $B_i$ respectively. 

The ARI evaluates the similarity between a set of assigned labels and the true labels in a range of $[-1,1]$, where $\text{ARI}<0$ means the clusters are worse than the results generated randomly and $\text{ARI}=1$ means the cluster is perfectly matched with known labels.

\section{Experimental Results}
\subsection{Novelty detection} 
Table\,\ref{table:nov_results} shows that our novelty detector yields the highest overall correction at $91\%$ and the highest recognition rate of identifying known objects as known objects at $95\,\%$. For detecting novel objects, our method could reach $89\,\%$ correction which is higher than Isolated forest and GMM results ($80\,\%$ and $84\,\%$ respectively), but it is lower than NNO results which is $96\,\%$. However, NNO only provides a $72\,\%$ accuracy for identifying known objects which is the lowest value. This shows that our method could maintain a relatively high recognition rate in both identifying known objects and detecting novel objects.

\begin{table}[h]
 \begin{center}
 \caption{Accuracy of novelty detection: distinction between known and novel objects}
\label{table:nov_results}
\begin{tabular}{lccc}
\hline
\multirow{2}{*}{\textbf{Methods}} & \multicolumn{3}{c}{\textbf{Accuracy (\%)}} \\
                                  & \textbf{Known}    & \textbf{Novel}    & \textbf{Overall}          \\ \hline
Isolated Forest                   & 88.19 $\pm$ 5.91          & 80.76 $\pm$ 9.93          & 82.79 $\pm$ 7.00            \\
GMM \cite{Abderrahmane2020}                               & 88.28 $\pm$ 6.21         & 84.69 $\pm$ 5.38           & 86.64 $\pm$ 3.55            \\
NNO \cite{Abhijit2015}                              & 72.29 $\pm$ 9.26         & \textbf{96.28 $\pm$ 2.50}           & 90.36 $\pm$ 3.27             \\ \hline
Our                               & \textbf{95.76 $\pm$ 1.87}         & 89.21 $\pm$ 4.76            & \textbf{91.06 $\pm$ 2.98}   \\ \hline
\end{tabular}
\end{center}
\end{table}

\subsection{Multiclass classification of known objects} 

The classification results of known objects are shown in Table\,\ref{table:class_results}. A recognition rate of $95\,\%$ was achieved when the Naive Bayes classifier was used to classify identified known objects from the novelty detector. The results were slightly higher than other classifiers such as NN and SVM which is around $94\,\%$ recognition rate. On the other hand, using other novelty detectors such as Isolated forest and GMM with the same Naive Bayes classifier provided recognition rates around $85-88\,\%$. Lastly, NNO yielded a $91\,\%$ recognition rate. These results show that the Naive Bayes classifier provided the highest recognition.

\begin{table}[h]
 \begin{center}
 \caption{Recognition rate for multiclass classification of known objects}
\label{table:class_results}
\begin{tabular}{lc}
\hline
\textbf{Method}             & \textbf{Recognition rate (\%)} \\ \hline
Isolated forest+NB & 85.73 $\pm$ 6.41       \\
GMM+NB             & 88.13 $\pm$ 4.38      \\
NNO \cite{Abhijit2015}             & 91.36 $\pm$ 3.41      \\ \hline
Our, but NN        & 94.16 $\pm$ 1.85      \\
Our, but SVM       & 93.78 $\pm$ 1.96      \\ \hline
Our                & \textbf{95.58 $\pm$ 1.61}      \\ \hline

\end{tabular}
\end{center}
\end{table}

\subsection{Clustering of novel objects}
Table\,\ref{table:cluster_results} shows that our clustering algorithm could achieve values of ARI at $0.70$. The K-means-based algorithm yields values of $0.60-0.64$ ARI. On the other hand, the DBscan-based algorithm provided a lower ARI at $0.64-0.66$. Using VAE as a new cluster parameters estimator resulted in the lowest value of ARI at almost $0$. This result is even lower than the framework that randomly generates cluster parameters. These results show that our clustering algorithm outperforms other methods. Besides, using a regression model to estimate cluster parameters yields better results than using random parameters and VAE.

\begin{table}[h]
 \begin{center}
 \caption{Evaluation of clustering results for novel objects}
\label{table:cluster_results}
\begin{tabular}{lc}
\hline
\textbf{Method}           & \multicolumn{1}{c}{\textbf{ARI}}             \\ \hline
K-means  \         & 0.642 $\pm$ 0.123  \\
IOKM  \cite{ABERNATHY2022}  & 0.604 $\pm$ 0.078  \\
DBscan         & 0.667 $\pm$ 0.135  \\

DenStream \cite{Cao2006}      & 0.646 $\pm$ 0.082  \\ \hline
Our, but random   & 0.120 $\pm$ 0.025 \\
Our, but VAE    & 0.001 $\pm$ 0.002 \\ \hline
Our              &  \textbf{0.701 $\pm$ 0.096} \\ \hline
\end{tabular}
\end{center}
\end{table}

\subsection{Clustering results with various novel classes threshold}

We investigated the performance of our clustering algorithms when the number of novel classes varied from 10 to 90\% of the total classes in the testing set. We also compared our results with the DenStream results where both algorithms are online algorithms and do not require the number of clusters. 

\begin{figure}[!bt]
\centering
\includegraphics[width=0.975\columnwidth]{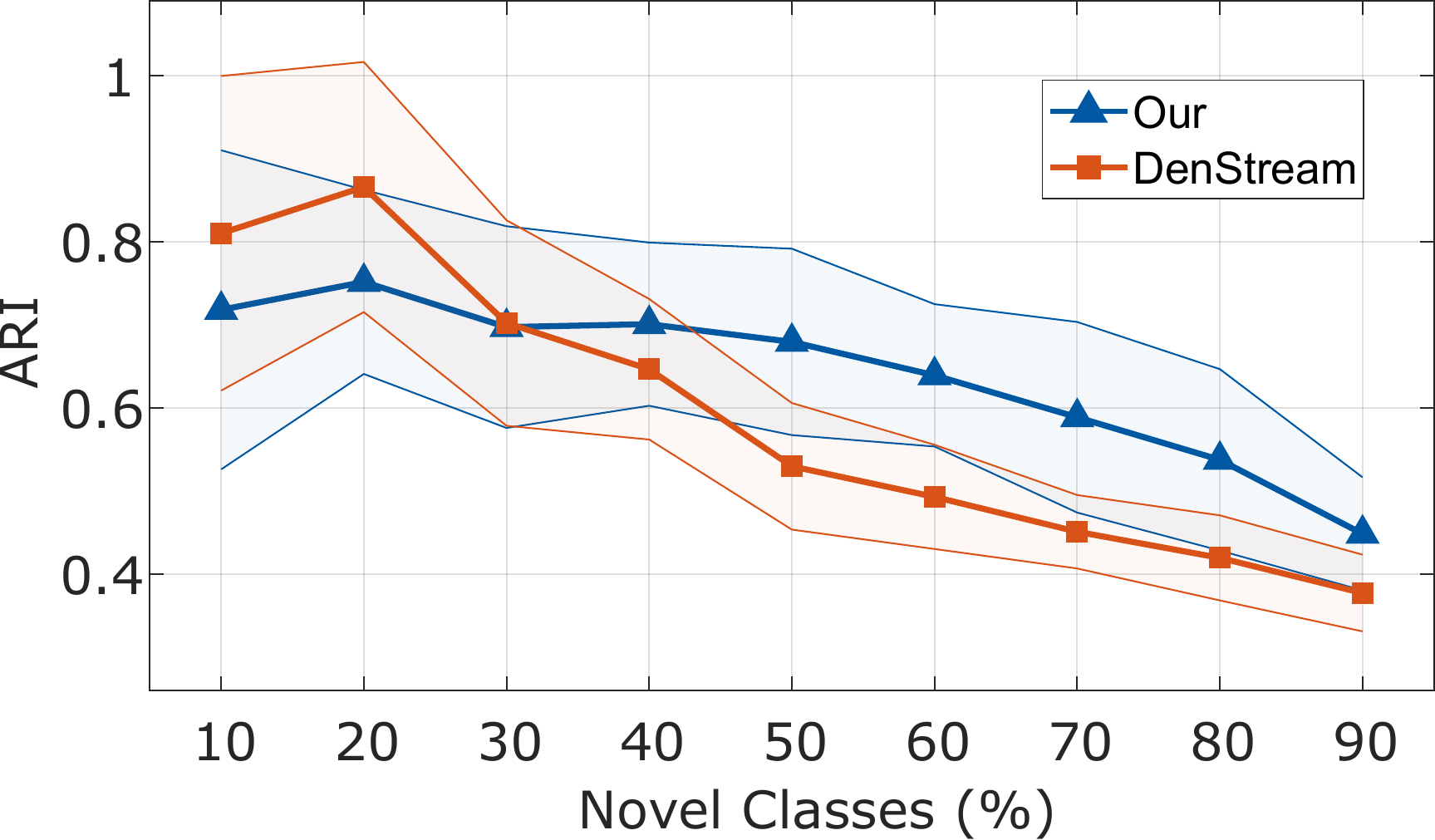}
\caption{Performance’s comparison of different algorithms to recognise objects with different posterior thresholds showing mean values of ARI and their standard deviation.}
\label{fig:results_num_unseen}
\end{figure}

Fig.\,\ref{fig:results_num_unseen} shows that mean values of the  ARI in both algorithms expectedly decreased with more novel classes in the testing set. Our clustering algorithm could achieve an ARI of 0.72 at 10\,\% novel classes and gradually decrease to 0.45 at 90\,\% novel classes. On the other hand, DenStream provided an ARI of 0.81 at 10\,\% of novel classes before the value decreased to 0.37 at 90\,\% of novel classes. 

The results demonstrated that our algorithm is less sensitive than the DesnStream when the number of novel classes is increased. In fact, our algorithm provides higher values of ARI than the DenStream in 40-90\,\% of novel classes. Both algorithms provide similar values of ARI which is 0.7 at 30\,\% novel classes. However, DenStream provides higher values of ARI than our results in the range of the novel classes 10-20\,\%.

\section{Hyperparameters Studies}
To investigate the effect of hyperparameters on clustering results, the values of $\alpha$, $\beta$, $N_{gen}$ in the cluster estimator and the updating threshold $\tau_{update}$ were varied. Each hyperparameter is evaluated with the 40\,\% of novel classes in the testing set and is reported by calculating the ARI using eq.\,\ref{eq:ARI}. 

\subsection{Hyperparameters in the new cluster estimator}

\begin{figure}[!bt]
\centering

\includegraphics[width=0.95\columnwidth]{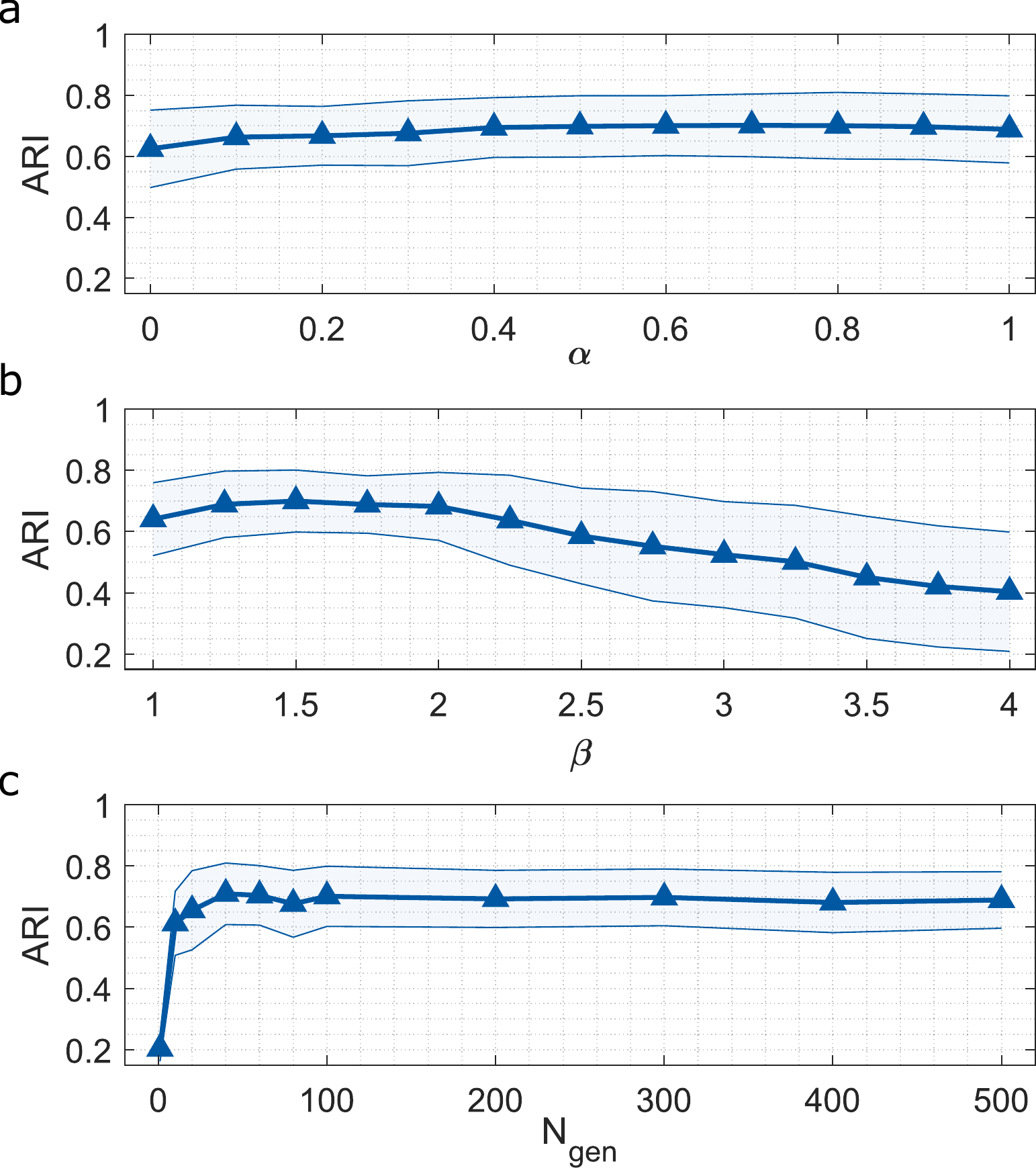}
\caption{The mean values of ARI and its standard deviation obtained using different values of (a) $\alpha$, (b) $\beta$, and (c) $N_{gen}$ in the new cluster estimator.}
\label{fig:results_hyperpara}
\end{figure}

The first hyperparameters in the new cluster estimator $\alpha$ control the location of the new cluster centre. 
Fig.\,\ref{fig:results_hyperpara}\,a shows that the mean value of the ARI increases from 0.62 at $\alpha=0$ to 0.70 at $\alpha=0.4$. Following this, all values of $\alpha$ above 0.4 have shown similar performance. These results suggest that the predicted mean from the new cluster estimator may not be accurate enough to use as the centre of the new cluster. However, shifting it to be close to the test sample helps improve the cluster results as the test sample is obtained from the real objects. 

The $\beta$ controls the standard deviation of datapoints inside the new cluster. Fig.\,\ref{fig:results_hyperpara}\,b shows mean values of the ARI obtained when the $\beta$ is varied from 1 to 4. Changes in the ARI are observed as the ARI is 0.64 at  $\beta=1$ and then the value increases until reaching its peak which is 0.70 at $\beta=1.5$. This is followed by the reduction of ARI as the $beta$ increases. These results suggest $\beta$ needs to be sufficiently large enough for test datapoints to lie in the correct cluster, but not too large that all datapoints lie in the same cluster.

Lastly, $N_{gen}$ defines the number of datapoints used to find the boundary parameter $\eta_c$. Fig.\,\ref{fig:results_hyperpara}\,c shows that the ARI increases from 0.20 at $N_{gen}=1$ to reach its saturated point which is 0.70 at $N_{gen}=40$. These results suggest that a small $N_{gen}$ may generate datapoints that are close to the cluster centre, thus the calculated boundary may be smaller than it is supported to be. A larger $N_{gen}$ may guarantee the correct boundary of the new cluster. However, too large $N_{gen}$ can cause a higher computational cost as more datapoints need to be calculated.

\subsection{Updating threshold for novel object clusters}
\begin{figure}[!bt]
\centering
\includegraphics[width=0.95\columnwidth]{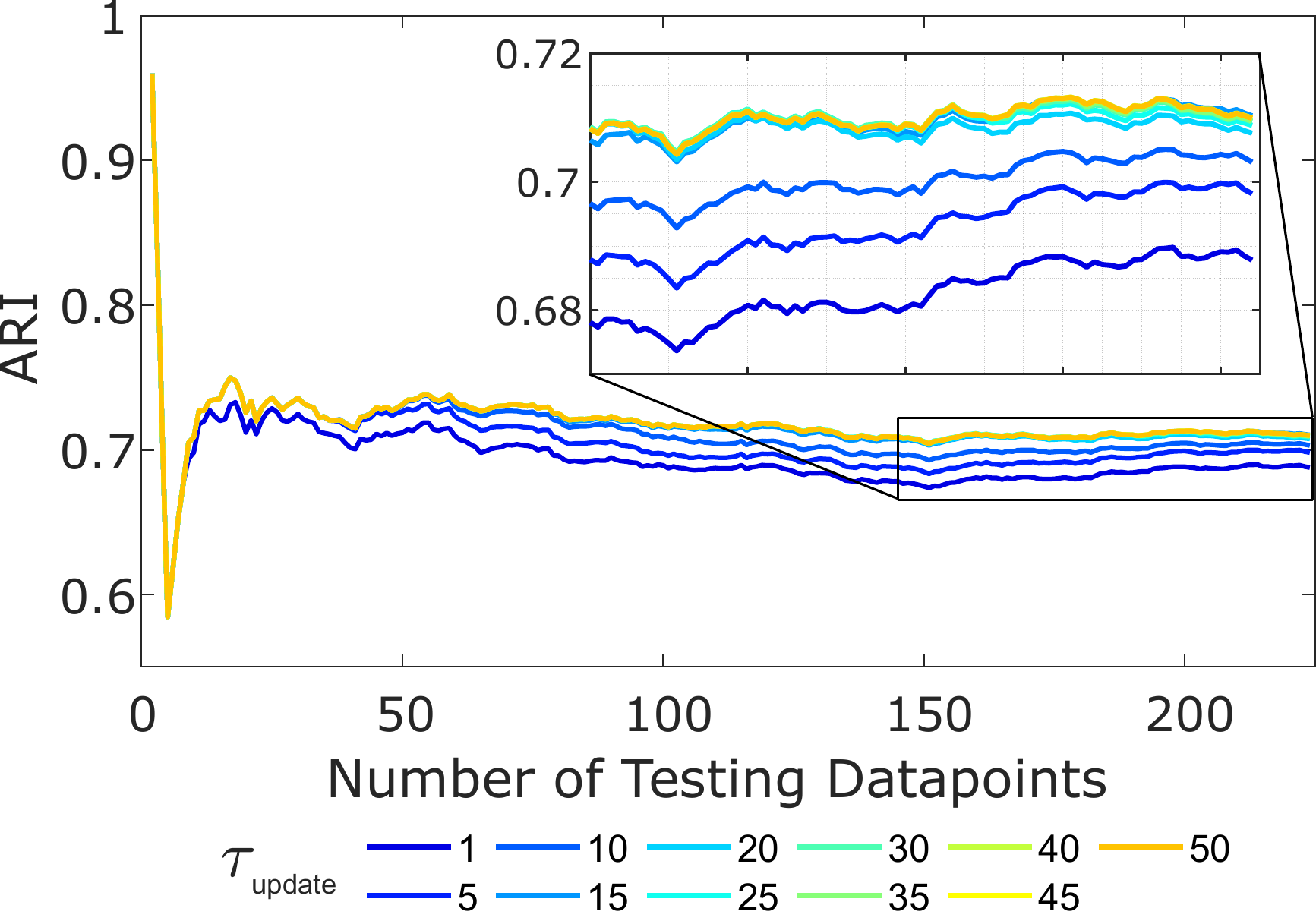}
\caption{Performance's comparison of different values of $\tau_{update}$ showing the evolution of the mean values of ARI as a function of the number of testing datapoints.}
\label{fig:results_numupdate}
\end{figure}

The $\tau_{update}$ defines when the new cluster parameters should be updated as the new datapoints are assigned to the cluster. Fig.\,\ref{fig:results_numupdate} illustrates that the values of ARI change from 0.96 to 0.6 when 5 new datapoints are fed at the beginning. However, the ARI starts to increase as more datapoints are added to the system until it reaches its saturation point depending on values of $\tau_{update}$. 

ARI yield values of 0.68-0.69 in the range of 1 to 10 for $\tau_{update}$, especially in the latter number of fed datapoints. These results suggest that updating parameters frequently could lead to slightly poor clustering results. When $\tau_{update}$ increases, the ARI also slightly increases. However, the ARI reaches its saturation at 0.71 after $\tau_{update}$ reaches 15. These results suggest that updating parameters when there are 15 new datapoints in the cluster is the balanced choice of update for this dataset.

\section{Discussion}
This paper introduced an open-set recognition framework based on the estimation of mechanical properties. This method extends the recognition framework in \cite{Uttayopas2023} to identify known objects and also to incrementally learn novel objects online. The algorithm's viability was demonstrated in simulation using the mechanical properties estimated from real objects. Our results show that it can detect novel objects with a 91\% accuracy rate and achieve a 95\% recognition rate for known objects. Importantly, it can label novel objects with a 0.71 ARI according to the ground-truth labels.

Our classification result is better than the alternative methods. Although the same Navie Bayes classifier is used to classify known objects detected by isolated forests and GMM, their results are lower than ours. Besides, NNO also provide poorer classification results as it cannot detect known object very well. These results suggest that novelty detectors have an influence on recognition rates, as failure to detect both known and novel objects can result in misclassification. On the other hand, using different classifiers with the same novelty detector yields similar classification results, illustrating the robustness of using mechanical properties as features.


The proposed clustering algorithm achieved the best performance because of the help of the new cluster parameters estimator. Unlike DenStream \cite{Cao2006}, our method exploits the knowledge of known objects to synthetically create a new cluster when encountering a novel object and update the new cluster when there are enough datapoints in it. This process aligns the cluster's size and position to be close to the ground truth, thus better clustering performance. Besides, the proposed method doesn't need a complete set of data like K-means and DBscan to perform clustering or a predefined number of clusters like K-mean and IOKM \cite{ABERNATHY2022}, which is beneficial for robots learning about new objects in real-world scenarios.


The ARI value with simple regression models as cluster parameter estimators is higher than with randomly generated parameters illustrating the benefit of using the regression model to exploit the known object knowledge to estimate the parameters of the new cluster. On the other hand, using VAE as the estimator results in the lowest values of ARI even though VAE has been successfully used to generate fake samples in visual-based zero-shot learning \cite{Verma2018}. This might be the fact that such generative methods require a large number of training samples to learn \cite{ABDERRAHMANE2018, Abderrahmane2020}. In our case, less than 25 trials for each object are in the training set which may cause the VAE model to be underfitting.

While the presented system could successfully recognise known objects and learn to cluster novel objects, some limitations should be acknowledged. First, the current system hyperparameters were tuned by grid search which requires a lot of computational time. This could be addressed by using more efficient optimisation methods. Second, the current system relies only on the initial training data, which can lead to poor parameter estimation for novel classes. To address this, the system should incorporate novel objects in a few-shot learning manner. However, incorrect initial labeling of the novel objects may result in poorer performance, thus more accurate clustering algorithm is needed.

In sum, the experiment results illustrate the potential of the proposed framework using mechanical properties to recognise known objects and incrementally learn novel objects to label them. This is done by first detecting known and novel objects. While the known objects are classified, the novel objects are used to synthetically create clusters awaiting more samples to ensure. It is noteworthy that using knowledge of known objects is useful to help create such clusters rather than without. While the framework successfully recognise and labels objects with various mechanical properties, considering the shape of objects and expanding the framework to be able to manipulate them would be useful in haptic exploration.

\printbibliography[]

@article{Luo2017Mechatronics,
title = "Robotic tactile perception of object properties: A review",
journal = "Mechatronics",
volume = "48",
pages = "54 - 67",
year = "2017",
issn = "0957-4158",
doi = "https://doi.org/10.1016/j.mechatronics.2017.11.002",
url = "http://www.sciencedirect.com/science/article/pii/S0957415817301575",
author = "S. Luo and Joao Bimbo and Ravinder Dahiya and Hongbin Liu",
}

@article{Dallaire2014,
title = "Autonomous tactile perception: A combined improved sensing and Bayesian nonparametric approach",
journal = "Robotics and Autonomous Systems",
volume = "62",
number = "4",
pages = "422 - 435",
year = "2014",
issn = "0921-8890",
doi = "https://doi.org/10.1016/j.robot.2013.11.011",
url = "http://www.sciencedirect.com/science/article/pii/S0921889013002285",
author = "P. Dallaire and Philippe Giguère and Daniel Émond and Brahim Chaib-Draa"
}

@article{Campolo2014,
title = {H-Man: A planar, H-shape cabled differential robotic manipulandum for experiments on human motor control},
journal = {Journal of Neuroscience Methods},
volume = {235},
pages = {285-297},
year = {2014},
issn = {0165-0270},
doi = {https://doi.org/10.1016/j.jneumeth.2014.07.003},
url = {https://www.sciencedirect.com/science/article/pii/S0165027014002477},
author = {Domenico Campolo and Paolo Tommasino and Kumudu Gamage and Julius Klein and Charmayne M.L. Hughes and Lorenzo Masia}
}

@Article{Bednarek2021,
AUTHOR = {Bednarek, Michal and Kicki, Piotr and Bednarek, Jakub and Walas, Krzysztof},
TITLE = {Gaining a Sense of Touch Object Stiffness Estimation Using a Soft Gripper and Neural Networks},
JOURNAL = {Electronics},
VOLUME = {10},
YEAR = {2021},
NUMBER = {1},
ARTICLE-NUMBER = {96},
URL = {https://www.mdpi.com/2079-9292/10/1/96},
ISSN = {2079-9292},
DOI = {10.3390/electronics10010096}
}

@INPROCEEDINGS{Su2015,
  author={Su, Zhe and Hausman, Karol and Chebotar, Yevgen and Molchanov, Artem and Loeb, Gerald E. and Sukhatme, Gaurav S. and Schaal, Stefan},
  booktitle={IEEE-RAS International Conference on Humanoid Robots}, 
  title={Force estimation and slip detection/classification for grip control using a biomimetic tactile sensor}, 
  year={2015},
  volume={},
  number={},
  pages={297-303},
  doi={10.1109/HUMANOIDS.2015.7363558}}

@article{Ren2017,
author = {Ren, Anye and Qi, Chenkun and Gao, Feng and Zhao, Xianchao and Sun, Qiao},
year = {2017},
pages = {1-9},
title = {Contact Stiffness Identification with Delay and Structural Compensation for Hardware-in-the-Loop Contact Simulator},
volume = {86},
journal = {Journal of Intelligent and Robotic Systems},
doi = {10.1007/s10846-016-0432-2}
}

@book{Stronge2018, place={Cambridge}, edition={2}, title={Impact Mechanics}, DOI={10.1017/9781139050227}, publisher={Cambridge University Press}, author={Stronge, W. J.}, year={2018}}

@INPROCEEDINGS{Yane2022,
  author={Yane, Kazuki and Nozaki, Takahiro},
  booktitle={IEEE International Conference on Advanced Motion Control}, 
  title={Recognition of Environmental Impedance Configuration by Neural Network Using Time-Series Contact State Response}, 
  year={2022},
  volume={},
  number={},
  pages={426-431},
  doi={10.1109/AMC51637.2022.9729306}}

@ARTICLE{Uttayopas2023,
  author={Uttayopas, Pakorn and Cheng, Xiaoxiao and Eden, Jonathan and Burdet, Etienne},
  journal={IEEE Transactions on Haptics}, 
  title={Object Recognition Using Mechanical Impact, Viscoelasticity, and Surface Friction During Interaction}, 
  year={2023},
  volume={16},
  number={2},
  pages={251-260},
  doi={10.1109/TOH.2023.3267523}}

@article{Gauthier2021,
	doi = {10.1038/s41467-021-25801-2},
  
	url = {https://doi.org/10.1038%2Fs41467-021-25801-2},
  
	year = 2021,
	month = {sep},
  
	publisher = {Springer Science and Business Media {LLC}
},
  
	volume = {12},
  
	number = {1},
  
	author = {Daniel J. Gauthier and Erik Bollt and Aaron Griffith and Wendson A. S. Barbosa},
  
	title = {Next generation reservoir computing},
  
	journal = {Nature Communications}
}

@ARTICLE{Mensink2013,
  author={Mensink, Thomas and Verbeek, Jakob and Perronnin, Florent and Csurka, Gabriela},
  journal={IEEE Transactions on Pattern Analysis and Machine Intelligence}, 
  title={Distance-Based Image Classification: Generalizing to New Classes at Near-Zero Cost}, 
  year={2013},
  volume={35},
  number={11},
  pages={2624-2637},
  doi={10.1109/TPAMI.2013.83}}

@ARTICLE{Abderrahmane2020,
  author={Abderrahmane, Zineb and Ganesh, Gowrishankar and Crosnier, André and Cherubini, Andrea},
  journal={IEEE Transactions on Industrial Informatics}, 
  title={A Deep Learning Framework for Tactile Recognition of Known as Well as Novel Objects}, 
  year={2020},
  volume={16},
  number={1},
  pages={423-432},
  doi={10.1109/TII.2019.2898264}}

@INPROCEEDINGS{Abhijit2015,
  author={Bendale, Abhijit and Boult, Terrance},
  booktitle={2015 IEEE Conference on Computer Vision and Pattern Recognition (CVPR)}, 
  title={Towards Open World Recognition}, 
  year={2015},
  volume={},
  number={},
  pages={1893-1902},
  doi={10.1109/CVPR.2015.7298799}}

@INPROCEEDINGS{Cao2006,
author = {Feng Cao and Martin Estert and Weining Qian and Aoying Zhou},
title = {Density-Based Clustering over an Evolving Data Stream with Noise},
booktitle = {SIAM International Conference on Data Mining},
year = {2006},
chapter = {},
pages = {328-339},
doi = {10.1137/1.9781611972764.29},

}

@article{ABERNATHY2022,
title = {The incremental online k-means clustering algorithm and its application to color quantization},
journal = {Expert Systems with Applications},
volume = {207},
pages = {117927},
year = {2022},
issn = {0957-4174},
doi = {https://doi.org/10.1016/j.eswa.2022.117927},
author = {Amber Abernathy and M. Emre Celebi},
}

@inproceedings{cao22,
  title={Open-World Semi-Supervised Learning},
  author={Cao, Kaidi and Brbi\'c, Maria and Leskovec, Jure},
  booktitle={International Conference on Learning Representations (ICLR)},
  year={2022},
}

@INPROCEEDINGS{Liu2023,
  author={Liu, Kunhong and Yang, Qianhui and Xie, Yu and Huang, Xiangyi},
  booktitle={2023 IEEE International Conference on Robotics and Automation (ICRA)}, 
  title={Towards Open-Set Material Recognition using Robot Tactile Sensing}, 
  year={2023},
  volume={},
  number={},
  pages={10345-10351},
  doi={10.1109/ICRA48891.2023.10161108}}

@article{ABDERRAHMANE2018,
title = {Haptic Zero-Shot Learning: Recognition of objects never touched before},
journal = {Robotics and Autonomous Systems},
volume = {105},
pages = {11-25},
year = {2018},
issn = {0921-8890},
doi = {https://doi.org/10.1016/j.robot.2018.03.002},
url = {https://www.sciencedirect.com/science/article/pii/S0921889017307492},
author = {Zineb Abderrahmane and Gowrishankar Ganesh and André Crosnier and Andrea Cherubini},
}

@INPROCEEDINGS {Verma2018,
author = {V. Verma and G. Arora and A. Mishra and P. Rai},
booktitle = {2018 IEEE/CVF Conference on Computer Vision and Pattern Recognition (CVPR)},
title = {Generalized Zero-Shot Learning via Synthesized Examples},
year = {2018},
volume = {},
issn = {},
pages = {4281-4289},
doi = {10.1109/CVPR.2018.00450},
publisher = {IEEE Computer Society},
address = {Los Alamitos, CA, USA},
month = {jun}
}

@INPROCEEDINGS{Cao2020,
  author={Cao, Guanqun and Zhou, Yi and Bollegala, Danushka and Luo, Shan},
  booktitle={2020 IEEE/RSJ International Conference on Intelligent Robots and Systems (IROS)}, 
  title={Spatio-temporal Attention Model for Tactile Texture Recognition}, 
  year={2020},
  volume={},
  number={},
  pages={9896-9902},
  doi={10.1109/IROS45743.2020.9341333}}

@INPROCEEDINGS{Taunyazov2019,
  author={Taunyazov, Tasbolat and Koh, Hui Fang and Wu, Yan and Cai, Caixia and Soh, Harold},
  booktitle={2019 International Conference on Robotics and Automation (ICRA)}, 
  title={Towards Effective Tactile Identification of Textures using a Hybrid Touch Approach}, 
  year={2019},
  volume={},
  number={},
  pages={4269-4275},
  doi={10.1109/ICRA.2019.8793967}}

@ARTICLE{Bhattacharjee2018,
  author={Bhattacharjee, Tapomayukh and Clever, Henry M. and Wade, Joshua and Kemp, Charles C.},
  journal={IEEE Robotics and Automation Letters}, 
  title={Multimodal Tactile Perception of Objects in a Real Home}, 
  year={2018},
  volume={3},
  number={3},
  pages={2523-2530},
  doi={10.1109/LRA.2018.2810956}}

@inproceedings{Wu2022,
author = {Wu, Zhiheng and Lu, Yue and Chen, Xingyu and Wu, Zhengxing and Kang, Liwen and Yu, Junzhi},
title = {UC-OWOD: Unknown-Classified Open World Object Detection},
year = {2022},
isbn = {978-3-031-20079-3},
publisher = {Springer-Verlag},
address = {Berlin, Heidelberg},
url = {https://doi.org/10.1007/978-3-031-20080-9_12},
doi = {10.1007/978-3-031-20080-9_12},
booktitle = {Computer Vision – ECCV 2022: 17th European Conference, Tel Aviv, Israel, October 23–27, 2022, Proceedings, Part X},
pages = {193–210},
numpages = {18},
location = {Tel Aviv, Israel}
}

@inproceedings{Ma2022,
author = {Ma, Zeyu and Yang, Yang and Wang, Guoqing and Xu, Xing and Shen, Heng Tao and Zhang, Mingxing},
title = {Rethinking Open-World Object Detection in Autonomous Driving Scenarios},
year = {2022},
isbn = {9781450392037},
publisher = {Association for Computing Machinery},
address = {New York, NY, USA},
url = {https://doi.org/10.1145/3503161.3548165},
doi = {10.1145/3503161.3548165},
booktitle = {Proceedings of the 30th ACM International Conference on Multimedia},
pages = {1279–1288},
numpages = {10},
location = {Lisboa, Portugal},
series = {MM '22}
}

@INPROCEEDINGS {Bogdoll2022,
author = {D. Bogdoll and M. Nitsche and J. Zollner},
booktitle = {2022 IEEE/CVF Conference on Computer Vision and Pattern Recognition Workshops (CVPRW)},
title = {Anomaly Detection in Autonomous Driving: A Survey},
year = {2022},
volume = {},
issn = {},
pages = {4487-4498},
doi = {10.1109/CVPRW56347.2022.00495},
url = {https://doi.ieeecomputersociety.org/10.1109/CVPRW56347.2022.00495},
publisher = {IEEE Computer Society},
address = {Los Alamitos, CA, USA},
month = {jun}
}

@InProceedings{Zheng2022,
    author    = {Zheng, Jiyang and Li, Weihao and Hong, Jie and Petersson, Lars and Barnes, Nick},
    title     = {Towards Open-Set Object Detection and Discovery},
    booktitle = {Proceedings of the IEEE/CVF Conference on Computer Vision and Pattern Recognition (CVPR) Workshops},
    month     = {June},
    year      = {2022},
    pages     = {3961-3970}
}

@INPROCEEDINGS{Vareto2017,
  author={Vareto, Rafael and Silva, Samira and Costa, Filipe and Schwartz, William Robson},
  booktitle={2017 IEEE International Joint Conference on Biometrics (IJCB)}, 
  title={Towards open-set face recognition using hashing functions}, 
  year={2017},
  volume={},
  number={},
  pages={634-641},
  doi={10.1109/BTAS.2017.8272751}}

@INPROCEEDINGS{Gunther2017,
  author={Günther, M. and Hu, P. and Herrmann, C. and Chan, C. H. and Jiang, M. and Yang, S. and Dhamija, A. R. and Ramanan, D. and Beyerer, J. and Kittler, J. and Jazaery, M. Al and Nouyed, M. I. and Guo, G. and Stankiewicz, C. and Boult, T. E.},
  booktitle={2017 IEEE International Joint Conference on Biometrics (IJCB)}, 
  title={Unconstrained Face Detection and Open-Set Face Recognition Challenge}, 
  year={2017},
  volume={},
  number={},
  pages={697-706},
  doi={10.1109/BTAS.2017.8272759}}

@article{Boult2019, 
title={Learning and the Unknown: Surveying Steps toward Open World Recognition}, 
volume={33}, 
url={https://ojs.aaai.org/index.php/AAAI/article/view/5054}, 
DOI={10.1609/aaai.v33i01.33019801}, 
number={1}, 
journal={Proceedings of the AAAI Conference on Artificial Intelligence}, 
author={Boult, T. E. and Cruz, S. and Dhamija, A.R. and Gunther, M. and Henrydoss, J. and Scheirer, W.J.}, 
year={2019}, 
month={Jul.}, 
pages={9801-9807} }

@INPROCEEDINGS{Salomon2020,
  author={Salomon, Gabriel and Britto, Alceu and Vareto, Rafael H. and Schwartz, William R. and Menotti, David},
  booktitle={2020 International Conference on Systems, Signals and Image Processing (IWSSIP)}, 
  title={Open-set Face Recognition for Small Galleries Using Siamese Networks}, 
  year={2020},
  volume={},
  number={},
  pages={161-166},
  doi={10.1109/IWSSIP48289.2020.9145245}}

@InProceedings{Joseph2021,
    author    = {Joseph, K J and Khan, Salman and Khan, Fahad Shahbaz and Balasubramanian, Vineeth N},
    title     = {Towards Open World Object Detection},
    booktitle = {Proceedings of the IEEE/CVF Conference on Computer Vision and Pattern Recognition (CVPR)},
    month     = {June},
    year      = {2021},
    pages     = {5830-5840}
}

@InProceedings{Yang2018,
author = {Yang, Hong-Ming and Zhang, Xu-Yao and Yin, Fei and Liu, Cheng-Lin},
title = {Robust Classification With Convolutional Prototype Learning},
booktitle = {Proceedings of the IEEE Conference on Computer Vision and Pattern Recognition (CVPR)},
month = {June},
year = {2018}
}

@InProceedings{Verma2017,
author="Verma, Vinay Kumar
and Rai, Piyush",
editor="Ceci, Michelangelo
and Hollm{\'e}n, Jaakko
and Todorovski, Ljup{\v{c}}o
and Vens, Celine
and D{\v{z}}eroski, Sa{\v{s}}o",
title="A Simple Exponential Family Framework for Zero-Shot Learning",
booktitle="Machine Learning and Knowledge Discovery in Databases",
year="2017",
publisher="Springer International Publishing",
address="Cham",
pages="792--808",
isbn="978-3-319-71246-8"
}


\end{document}